%% file: main.tex
\documentclass[graybox]{svmult}

\usepackage{mathptmx}       
\usepackage{helvet}         
\usepackage{courier}        
\usepackage{type1cm}        
\usepackage{makeidx}         
\usepackage{graphicx}        
\usepackage{multicol}        
\usepackage[bottom]{footmisc}
\usepackage{listings}
\usepackage{verbatim}

\makeindex             

\begin{document}

\title*{Semantic CPPS in Industry 4.0}

\author{Giuseppe Fenza and Mariacristina Gallo and Vincenzo Loia and Domenico Marino and Francesco Orciuoli and Alberto Volpe}
\authorrunning{Fenza G., Gallo M., Loia V., Marino D., Orciuoli F., Volpe A.} 
\institute{Giuseppe Fenza, Mariacristina Gallo, Vincenzo Loia, Domenico Marino, Francesco Orciuoli, \\Alberto Volpe \at Dipartimento di Scienze Aziendali-Management and Innovation Systems,\\University of Salerno, Fisciano, Italy\\
\email{gfenza@unisa.it, mgallo@unisa.it, loia@unisa.it, domenicomarino42@gmail.com,\\ forciuoli@unisa.it, volpea.93@icloud.com}
}

\maketitle

\abstract*{}

\abstract{
Cyber-Physical Systems (CPS) play a crucial role in the era of the 4th Industrial Revolution. Recently, the application of the CPS to industrial manufacturing leads to a specialization of them referred as Cyber-Physical Production Systems (CPPS).
Among other challenges, CPS and CPPS should be able to address interoperability issues, since one of their intrinsic requirement is the capability to interface and cooperate with other systems. On the other hand, to fully realize the Industry 4.0 vision, it is required to address horizontal, vertical, and end-to-end integration enabling a complete awareness through the entire supply chain. 
In this context, Semantic Web standards and technologies may have a promising role to represent manufacturing knowledge in a machine-interpretable way for enabling communications among heterogeneous Industrial assets. This paper proposes an integration of Semantic Web models available at state of the art for implementing a 5C architecture mainly targeted to collect and process semantic data stream in a way that would unlock the potentiality of data yield in a smart manufacturing environment. 
The analysis of key industrial ontologies and semantic technologies allows us to instantiate an example scenario for monitoring Overall Equipment Effectiveness (OEE). The solution uses the SOSA ontology for representing the semantic data stream. Then, 
C-SPARQL queries are defined for periodically carrying out useful KPIs to address the proposed aim. \footnote{This is a post-peer-review, pre-copyedit version of an article published Barolli L., Amato F., Moscato F., Enokido T., Takizawa M. (eds) Advanced Information Networking and Applications. AINA 2020. Advances in Intelligent Systems and Computing, vol 1151. Springer, Cham. The final authenticated version is available online at: https://doi.org/10.1007/978-3-030-44041-1\_91}
}

\section{Introduction}
\label{sec:1}
Cyber-Physical Systems (CPS) are a combination of physics with cyber components potentially networked and tightly interconnected. The physics components are \textbf{Plants}, \textbf{Processes}, and \textbf{Systems}, while the cyber components are \textbf{Computation}, \textbf{Software}, and \textbf{Code}. These systems are networked and interconnected thanks to interfaces that convert analog to a digital signal and vice-versa. In this way, each physic component has its digital twins that work in symbiosis to perform efficiently, 
enabling knowledge sharing and fast decision making. 

Recently, the 4th Industrial Revolution and the relevance of the CPS inside the manufacturing plan lead to the definition of a niche of CPS applied to smart manufacturing, named Cyber-Physical Production Systems (CPPS). 

Due to the heterogeneity of integrated components, the interoperability inside the CPPS begins a crucial requirement. It  enables the system
to reach and exploit data produced by other systems, and so on. However, a system of systems built with interoperability in mind would be difficult to manage in the long run and will slow the business processes over time. 
In other words, it becomes an obstacle to shareable knowledge opportunities for the systems and the stakeholders of the business.

Since a continuous flow of time-sensitive information is necessary for the plant activities, another relevant requirement of the CPPs is the ability to manage streams of data points. These data points are characterized by a time dimension that certifies the state of the producing source in a specific time window. It follows that procedures and analytics should be time-aware in terms of the processed stream. They should produce reports that resume the state in a time window.

In this paper, we propose a semantic component for the Cognition level of the usual 5C architecture (see Section~\ref{5c}). 
The semantic component is used to model the entire business, from the production floor to the management level. Each system models specific data points leveraging internal and external 
knowledge (i.e., produced by other systems). 
The idea is using the SOSA ontology to represent the different levels of the business 
to enable reasoning and unlock actionable knowledge of heterogeneous information sources \cite{3}. The use of ontologies constitutes a powerful solution to capture and to share collective knowledge between stakeholders \cite{article}\cite{hoppe2017shifting}. 
Furthermore, by applying a stream reasoning (i.e., C-SPARQL\footnote{http://streamreasoning.org/resources/c-sparql}), we enable a mechanism of continuous queries in order to retrieve specific knowledge inside the time windows.

The remaining of the document is structured as follows. The next section discusses related works that apply semantic models to CPPS. Section \ref{tech} describes the available semantic technologies, while Section \ref{approach} details our approach. A sample scenario is presented in Section \ref{scenario} that proves the usefulness of the proposed component applied to the manufacturing domain. Finally, Section \ref{conclusion} ends the paper.

\section{Related Works}
\label{sec:2}
\input{rw.tex}
\section{Semantic Technologies for Web of Things}
\label{tech}
\input{tech.tex}

\section{Our Approach}\label{approach}
A usual smart manufacturing environment is highly automated and networked. Each station is strictly interlinked with a digital twin to collaborate for performing reliably and being aware of the context in which they are placed. 
In this section, it will be provided a brief description of the 5C architecture that represents the fundamentals stone on which our approach (discussed in the next section) has been developed.

\subsection{5C Architecture}\label{5c}

\begin{figure}
    \centering
  \includegraphics[width=0.8\linewidth]{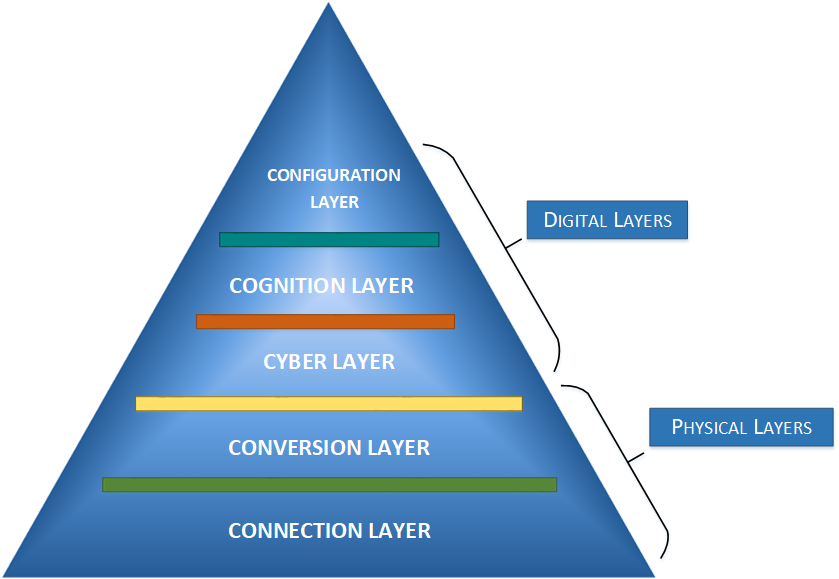}
  \caption{5C architecture of a Cyber-Physical System.}
  \label{fig:5c}
\end{figure}

In the area of Industrial Internet, 5C architecture provides a step-by-step guide for developing and deploying a CPS \cite{li2017industrial}. 
The Industrial Internet, viewed as the application of CPS \cite{drath2014industrie},  is composed of five layers: smart connection level, data-to-information conversion level, cyber level, cognition level, and configuration level (see Figure \ref{fig:5c}).

The Connection layer has the goal of managing data acquisition from the sensors and controllers based on heterogeneous hardware and software platforms. This complex task can be achieved thanks to the use of the protocols capable of handling different hardware and software. 

The Conversion layer uses the previous layer's output and transforms data into information thanks to the support of external data sources. This component is crucial to realize self-aware physical components.

The Cyber layer is the information hub for data processing. It groups the output of the previous layer represented by individual information to support decision making done on the upper layer. It analyzes and compares the current performance of the group of stations and predicts future performances.

The Cognition layer collects and integrates machinery information to provide knowledge of the whole production. In this layer, big data analytic pipelines and knowledge representation are predominant and helps the upper layer (the Configuration one) in managing the entire system and in providing relevant KPI related to the environment.

Then the last layer (i.e., Configuration) supports the managers and shows the decision taken by the Cognition layer with respect to the production floor. It acts as a smart control system to provide self-configuration and self-adaptiveness.

\subsection{Semantic Approach and Cognition Layer}
Micro, small, medium, and large systems have to be conceptualized with interoperability in mind: the entire ecosystem has to be able to reach data produced by other systems, and so on. 
This work emphasizes the need for a highly interoperable component enabling integration among different systems in the industrial asset, like Enterprise Resource Planning (ERP) and Manufacturing Execution System (MES). 

The solution proposes the adoption of semantic technologies inside the \emph{Cognitive Layer} of 5C architecture. The idea is to use the SOSA Ontology to model both static RDF data that represents the described companies’ assets and streaming part that is the on-line information that crosses the shop floor. Using the SOSA Ontology allows us to: (1) exploiting an ontology that already models fundamental concepts about sensors, samplings, procedures, and so on. (2) Reusing existing ontology models. One of the main benefits of this abstraction is the integration of, for instance, ERP and MES data allowing to support decision-making processes.

Through SOSA's classes and properties, we represent the entire environment and annotate raw data that came from the \emph{Cyber Layer}. Then, by adopting a reasoner, it is also possible to infer new facts based on the forwarded rules. In that way, the component creates knowledge that is the sum of facts, forward rules, backward rules, and goals. Annotated data are stored in a triplestore (e.g., RDF4J\footnote{https://rdf4j.org/}, AllegroGraph\footnote{https://franz.com/agraph/allegrograph/}, and so on.), offering a SPARQL endpoint for submitting queries. 
Such knowledge, inside the \emph{Configuration Layer}, supports managers in supervising the business processes. 

Sensors produce a rapidly evolving huge amount of data whose annotated version grows in turn. So, the semantic component must develop reliable production status in an online and proactive manner. To satisfy this requirement, the semantic component leverages C-SPARQL to do continuous queries over sensors' streams. By designing suitable queries, the component can support managers' decisions giving them focused knowledge for each specific time-frame. 

\section{Sample Scenario: Overall Equipment Effectiveness}
\label{scenario}
\begin{figure}
  \includegraphics[width=\linewidth]{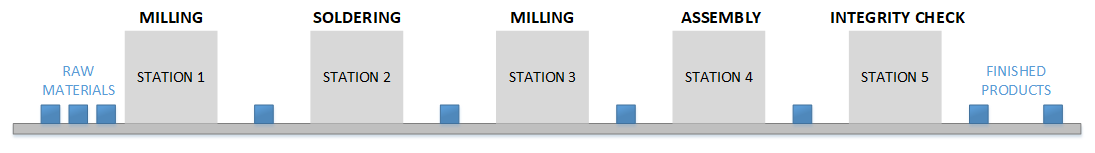}
  \caption{Basic production line with five stations.}
  \label{fig:prodLine}
\end{figure}

This section aims to describe an application scenario in which the defined semantic component could result useful. The objective is to support managers by generating up-to-date KPIs related to the production line showed in Figure \ref{fig:prodLine}.

\begin{figure}
  \includegraphics[width=\linewidth]{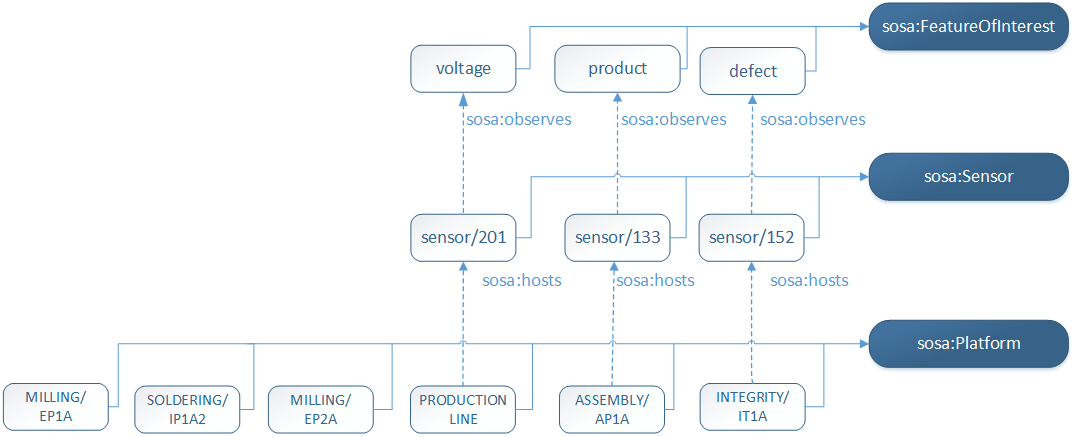}
  \caption{Basic production line modeled with SOSA ontology.}
  \label{fig:availability}
\end{figure}

The semantic component intends to provide a fast and accurate \emph{Synthetic Production Rate} that measures the efficiency of a facility. In particular, we will define the C-SPARQL queries able to evaluate: Availability, Efficiency of Performance, and Product Quality Rate of the factory. The starting point is the subset of SOSA ontology entities shown in Fig. \ref{fig:availability} that presents a basic model related to a production plan. Blue boxes indicate SOSA classes, and white boxes indicate class instances, while solid lines represent \emph{RDF:type} relations, and dashed lines, properties between entities.
In particular, Fig. \ref{fig:availability} shows the five stations that are instances of the SOSA class Platform, with an additional Platform instance for the production line. These platforms host sensors observing features of interest. Following we will discuss some sampling C-SPARQL queries related to this ontology schema. 

\subsection{Availability}
Availability is one of the three OEE factors, and its score takes into consideration any event (planned and unplanned) that stops the production. Planned stops refer to changes over time, while unplanned stops refer to equipment failures and material shortage. A lower availability score indicates problems or faults during the production; it leads to higher cycle-time that can influence the ability to satisfy demand. It is computed by recording the planned and unplanned stops. In particular, it is defined by Equation~\ref{availability}, where $TotalTime$ represents the full-time window (e.g., one day) while $DownTime$ is the sum of the planned and unplanned stops.
\begin{equation}\label{availability}
    {Availability} = \frac{TotalTime - DownTime}{TotalTime}
\end{equation}

In the following code blocks, two streams that register and compute the $DownTime$ and $Availability$ values, respectively, are defined.

\begin{lstlisting}[caption={DownTime evaluation stream},label={lst:downtime}, breaklines= true]
REGISTER STREAM DownTime COMPUTED EVERY 24h AS
  PREFIX sosa: <http://www.w3.org/ns/sosa/>
  SELECT ?downTime
  FROM STREAM <http://../production> [RANGE 24h STEP 1m]
  WHERE {?sensor sosa:observes ?voltage.
         ?voltage rdf:type sosa:FeatureOfInterest.
         ?productionLine sosa:hosts ?sensor}
  AGGREGATE {(?downTime, COUNT, {?voltage})
    FILTER (?voltage < 5 && ?productionLine = <http://.../ProductionLine>)}
\end{lstlisting}
\begin{lstlisting}[caption={Availability evaluation stream},label={lst:availability}]
REGISTER STREAM Availability COMPUTED EVERY 24h AS
  PREFIX sosa: <http://www.w3.org/ns/sosa/>
  SELECT (1440-?downTime)/1440 AS ?availability
  FROM STREAM <http://../DownTime> [RANGE 24h STEP 1m]
\end{lstlisting}

In Listing~\ref{lst:downtime}, we read voltage value from the specific sensor. We assume if such value is less then $5V$, the production line is down. So, the query counts the "down" triples and returns the total value. Successively, as expressed in Listing~\ref{lst:availability}, assuming that the $TotalTime$ is $1440$ (i.e., $24*60$), the query returns the resulting availability.

\subsection{Efficiency Of Performance}
The Efficiency of Performance (or simply performance) is a KPI that indicates how fast the process is running. It takes into consideration the ideal cycle time, the total production, and the operating time and is calculated as expressed in Equation~\ref{perform}.

\begin{equation}\label{perform}
    Efficiency Of Performance = \frac{Theoretical Cycle Time * Total Production}{Operating Time}
\end{equation}
The $Theoretical Cycle Time$ is specific for each production. It indicates the type of process, task sequence, technological dependencies, and duration of each task. The $Total Production$ regards the number of final products for the production line while the $Operating Time$ is the inverse of the $Down Time$ defined previously.

The evaluation of this KPI needs two different queries. The first one (see Listing~\ref{lst:total}) computes $TotalProduction$  by counting products passing through the assembly sensor. The second one evaluates the $Efficiency Of Performance$ by exploiting both $Total Production$ and $Down Time$ determined previously. For simplicity, the $TheoreticalCycleTime$ is supposed to be a constant (i.e., $25$ minutes). The corresponding query is described in Listing~\ref{lst:performance}.

\begin{lstlisting}[caption={TotalProduction evaluation stream},label={lst:total}]
REGISTER STREAM TotalProduction COMPUTED EVERY 24h AS
  PREFIX sosa: <http://www.w3.org/ns/sosa/>
  SELECT ?total
  FROM STREAM <http://../production> [RANGE 24h STEP 1m]
  WHERE {
         ?assemblySensor sosa:observes ?product.
         ?product rdf:type sosa:FeatureOfInterest.
         ?platform sosa:hosts ?assemblySensor
         }
  AGGREGATE {(?total, COUNT, {?product})
    FILTER (?platform = <http://.../ASSEMBLY/AP1A>)}
\end{lstlisting}

\begin{lstlisting}[caption={Performance evaluation stream},label={lst:performance}]
REGISTER STREAM Performance COMPUTED EVERY 24h AS
  PREFIX sosa: <http://www.w3.org/ns/sosa/>
  SELECT (25 * ?total)/(1440-?downTime) AS ?performance
  FROM STREAM <http://../TotalProduction> [RANGE 24h STEP 1m]
  FROM STREAM <http://../DownTime> [RANGE 24h STEP 1m]
\end{lstlisting}


\subsection{Product Quality Rate}
The quality score indicates how many products (or components) have been produced without flaws, and it is computed by Equation~\ref{quality}. The $Total Production$ is the same as the previous equation, while the $Defected Production$ is the number of defective pieces.
\begin{equation}\label{quality}
    {Product Quality Rate} = \frac{Total Production - Defected Production}{Total Production}
\end{equation}

In this case, as expressed in Listing~\ref{lst:quality}, it is necessary to read the $Total Production$ value from the previously generated stream (i.e., "TotalProduction") and count the number of defective products passing through the integrity sensor. 

\begin{lstlisting}[caption={Quality evaluation stream},label={lst:quality}]
REGISTER STREAM Quality COMPUTED EVERY 24h AS
  PREFIX sosa: <http://www.w3.org/ns/sosa/>
  SELECT ((?total - ?defectTotal)/?total) AS ?quality
  FROM STREAM <http://../TotalProduction> [RANGE 24h STEP 1m]
  FROM STREAM <http://../production> [RANGE 24h STEP 1m]
  WHERE {
         ?integritySensor sosa:observes ?defect.
         ?defect rdf:type sosa:FeatureOfInterest.
         ?platform sosa:hosts ?integritySensor
         }
  AGGREGATE {(?defectTotal, COUNT, {?defect})
        FILTER (?platform = <http://.../INTEGRITY/IT1A>)}
\end{lstlisting}


\subsection{Overall Equipment Effectiveness (OEE)}
The Overall Equipment Effectiveness (OEE) resumes the overall production state by considering 
the three evaluated factors: Availability, Performance, and Quality, through Equation~\ref{oee}. A 100\% score indicates that the plant is manufacturing perfect parts and in the fastest way possible without stops.
To compute the final OEE value, it is necessary to read from the three streams and compute the product of the values in the specific time-window (see Listing~\ref{lst:oee}).
\begin{equation}\label{oee}
    {OEE} = Availability * Performance * Quality
\end{equation}

\begin{lstlisting}[caption={OEE evaluation query},label={lst:oee}]
REGISTER QUERY OEE COMPUTED EVERY 24h AS
    SELECT (?availability * ?performance * ?quality) AS ?oee
    FROM STREAM <http://../Availability> [RANGE 24h STEP 1m]
    FROM STREAM <http://../Performance> [RANGE 24h STEP 1m]
    FROM STREAM <http://../Quality> [RANGE 24h STEP 1m]
\end{lstlisting}

\section{Conclusion}
\label{conclusion}
CPPS enables high efficiency and dynamics that are transversal through the manufacturing plants and the business units of a firm. To unlock this potentiality is required higher interoperability that sometimes can occur in a complex system in resource and maintenance needs. 
By taking into consideration these issues, this work proposes a semantic-based approach. 
We demonstrate how a semantic component could be beneficial to CPPS by providing interoperability and shareable knowledge. In particular, an analysis of the key industrial ontologies and semantic technologies is provided, and a common scenario is analyzed to show how powerful can be the proposed solution. Examples of queries relative to OEE show how effective is the use of a semantic component inside the Cognitive layer. In fact, the model supports the business and manufacturing processes and improves awareness about the overall manufacturing plant. In conclusion, the semantic component has resulted useful for:
\begin{itemize}
    \item \textbf{Internal interoperability}: integration between the static knowledge represented by the industrial assets and the dynamic knowledge produced by the IoT sensors;
    \item \textbf{External interoperability}: integration between multiple business and industrial assets;
    \item \textbf{Time-sensitive KPIs}: computation of relevant and on-time KPIs useful during the decision-making process. 
\end{itemize}

\begin{acknowledgement}
This research was partially supported by the ECSEL-JU under the program ECSEL-Innovation Actions-2018 (ECSEL-IA) for research project CPS4EU (ID-826276) research project in the area Cyber-Physical Systems. 
\end{acknowledgement}

\bibliographystyle{plain}
\bibliography{mybib}{}

\end{document}

%% file: rw.tex
This work proposes 
a 5C architecture adopting Semantic Web technologies and standards for 
unlocking the knowledge usually isolated in silos built and managed by different systems, for instance, like MES (Manufacturing Execution System), ERP (Enterprise Resource Planning), and so forth. The example scenario (see Section \ref{scenario}) focuses on the provisioning of useful KPIs to monitor the Overall Equipment Effectiveness using semantic data stream collected in a smart manufacturing environment. 

Several systems integrating Semantic Web with Web of Things there exist. Some of them are focused on the modeling of CPS' capabilities with Semantic Web technologies \cite{kumar2019ontologies, willner2017semantic, larsen2016integrated}, and some others use them to approach CPS' security aspects~\cite{bou2017cyber, hatzivasilis2017real}.



A research work most related to ours is 
\cite{ansari2019prima}, it introduces a model called PriMa that uses a semantic layer to learn and reason about data that comes from stations and sensors of the CPPS and to produce feedback to a Decision Support Dashboard used by the maintenance manager. 
Shcherbakov et al.~\cite{shcherbakov2020proactive} exploit the real-time requirements by describing a concept model for a proactive decision support system design specifically for the maintenance of CPS to optimize downtime. 

Different from state of the art, our approach proposes a model aiming to support integration at different levels and for different purposes in the CPS context. As also demonstrated by the application scenario, we construct a knowledge base shareable by systems and subsystems.

%% file: tech.tex
In this section, the description focuses on the identification of semantic technologies for the web of things that are naturally applicable to the smart manufacturing and Industry 4.0. 

Devices should be able to describe themselves and the data that they yield in a way that is understandable from both machines and humans. 
This may be done by equipping the devices and data with Semantic Web standards (or standard de-facto), ontological models, enabling to converge in the meaning
of various concepts that describe data and devices, and fully interoperability among them. The formalization of the knowledge graph takes place through the construction of ontologies that allow us to share common meanings applied to a domain of interest and describe concepts semantically, in our case, IoT devices in the industrial field. 

Semantic technologies substantially lie on the use of meta-models as: \emph{Resource Description Framework (RDF)}\footnote{https://www.w3.org/RDF/}, \emph{Resource Description Framework Schema (RDFS)}\footnote{https://www.w3.org/TR/2000/CR-rdf-schema-20000327/}, \emph{Web Ontology Language (OWL)}\footnote{https://www.w3.org/OWL/}. Defined by the World Wide Web Consortium (W3C), they allow to construct ontologies and populate them. 

Populated ontologies can then be queried by the RDF query language \emph{Simple Protocol and RDF Query Language (SPARQL)}\footnote{https://www.w3.org/TR/rdf-sparql-query/}.

During the design and modeling of an ontology for the IoT domain, it is necessary to know relevant aspects about devices such as the function to be performed, the environment in which they operate, the role and privileges within the operating context. Furthermore, it is relevant to reuse existing solutions and, eventually, extend them. In 2009, a W3C’s research selected and reviewed 17 existing ontologies related to the IoT domain. As a result, the consortium creates the Semantic Sensor Network (SSN)\footnote{https://www.w3.org/TR/vocab-ssn/} ontology~\cite{compton2012ssn}. 


The following sections will describe SSN and its evolution 
to 
model IoT streams and sensor data. 
\subsection{Semantic Sensor Network Ontology}
The Semantic Sensor Networks (SSN)\footnote{https://www.w3.org/2005/Incubator/ssn/ssnx/ssn} ontology is an OWL ontology that aims to describe the capacities and properties of the sensors, the act of detection, and the following observations. It 
was introduced by the W3C Semantic Sensor Networks Incubator group (the SSN-XG) held from March 2009 to
September 2010. 

As a choice of the working group, the ontology contains only the relevant concepts and relationships for the sensors, leaving the concepts related to other domains to be imported from other ontologies, where necessary. It can describe
the sensors, their accuracy, observations, and the methods used for detection. 
The SSN ontology can be viewed from four main perspectives:
\begin{description}
\item [$\bullet$] The sensor perspective, with a focus on what to capture and how;
\item [$\bullet$] Data or observation perspective, with a focus on observations and related metadata;
\item [$\bullet$] The system perspective, with a focus on sensor systems and deployment;
\item [$\bullet$] The perspective of features and properties, which focuses on what property to detect, or what observations have been made about a particular property.
\end{description}

\subsection{Sensor-Observation-Sampling-Actuator Ontology}
During the standardization of the Semantic Sensor Network by the W3C and OGC working groups, Spatial Data had the task of revisioning the SSN by using the experiences obtained during the development of the prior version. Due to scope and audience changes, and to overcome the lack of interoperability with other ontologies, SSN was rewritten entirely by using a modular design that enables horizontal and vertical segmentation. 

The new SSN, published as W3C Recommendation and OGC Standard, is the SOSA framework\footnote{https://www.w3.org/2015/spatial/wiki/SOSA\_Ontology}. This framework is composed of main concepts that enable interoperability with the entities, relationships, and activities involved in sensing, sampling, and actuation. Another relevant aspect of the SOSA ontology is the event-centric approach that enables interoperability with Schema.org and Prov-O models and, most importantly, with standard business processes procedures.

The sensor represents an entity that reacts to changes in the environment (e.g., black-out of a production line, temperature changes in a department). This detection produces the value of a property thanks to the (observation) procedure that was observing the specific property. Subsequent action performed due to the generated value is triggered thanks to an actuator. The last primary entity is the platform that hosts sensors and actuators.

%% file: main.bbl
\begin{thebibliography}{10}

\bibitem{ansari2019prima}
Fazel Ansari, Robert Glawar, and Tanja Nemeth.
\newblock Prima: a prescriptive maintenance model for cyber-physical production
  systems.
\newblock {\em International Journal of Computer Integrated Manufacturing},
  pages 1--22, 2019.

\bibitem{bou2017cyber}
Elias Bou-Harb, Walter Lucia, Nicola Forti, Sean Weerakkody, Nasir Ghani, and
  Bruno Sinopoli.
\newblock Cyber meets control: A novel federated approach for resilient cps
  leveraging real cyber threat intelligence.
\newblock {\em IEEE Communications Magazine}, 55(5):198--204, 2017.

\bibitem{compton2012ssn}
Michael Compton, Payam Barnaghi, Luis Bermudez, Ra{\'u}L Garc{\'\i}A-Castro,
  Oscar Corcho, Simon Cox, John Graybeal, Manfred Hauswirth, Cory Henson,
  Arthur Herzog, et~al.
\newblock The ssn ontology of the w3c semantic sensor network incubator group.
\newblock {\em Web semantics: science, services and agents on the World Wide
  Web}, 17:25--32, 2012.

\bibitem{drath2014industrie}
Rainer Drath and Alexander Horch.
\newblock Industrie 4.0: Hit or hype?[industry forum].
\newblock {\em IEEE industrial electronics magazine}, 8(2):56--58, 2014.

\bibitem{hatzivasilis2017real}
George Hatzivasilis, Ioannis Papaefstathiou, and Charalampos Manifavas.
\newblock Real-time management of railway cps secure administration of iot and
  cps infrastructure.
\newblock In {\em 2017 6th Mediterranean Conference on Embedded Computing
  (MECO)}, pages 1--4. IEEE, 2017.

\bibitem{hoppe2017shifting}
Tobias Hoppe, Harald Eisenmann, Alexander Viehl, and Oliver Bringmann.
\newblock Shifting from data handling to knowledge engineering in aerospace
  industry.
\newblock In {\em 2017 IEEE International Systems Engineering Symposium
  (ISSE)}, pages 1--6. IEEE, 2017.

\bibitem{kumar2019ontologies}
Veera Ragavan~Sampath Kumar, Alaa Khamis, Sandro Fiorini, Joel~Lu{\'\i}s
  Carbonera, Alberto~Olivares Alarcos, Maki Habib, Paulo Goncalves, Howard Li,
  and Joanna~Isabelle Olszewska.
\newblock Ontologies for industry 4.0.
\newblock {\em The Knowledge Engineering Review}, 34, 2019.

\bibitem{larsen2016integrated}
Peter~Gorm Larsen, John Fitzgerald, Jim Woodcock, Peter Fritzson, J{\"o}rg
  Brauer, Christian Kleijn, Thierry Lecomte, Markus Pfeil, Ole Green, Stylianos
  Basagiannis, et~al.
\newblock Integrated tool chain for model-based design of cyber-physical
  systems: The into-cps project.
\newblock In {\em 2016 2nd International Workshop on Modelling, Analysis, and
  Control of Complex CPS (CPS Data)}, pages 1--6. IEEE, 2016.

\bibitem{li2017industrial}
Jian-Qiang Li, F~Richard Yu, Genqiang Deng, Chengwen Luo, Zhong Ming, and Qiao
  Yan.
\newblock Industrial internet: A survey on the enabling technologies,
  applications, and challenges.
\newblock {\em IEEE Communications Surveys \& Tutorials}, 19(3):1504--1526,
  2017.

\bibitem{article}
Zhou Liandong and Wang Qifeng.
\newblock Knowledge discovery and modeling approach for manufacturing
  enterprises.
\newblock {\em Intelligent Information Technology Applications, 2007 Workshop
  on}, 1:291--294, 01 2009.

\bibitem{shcherbakov2020proactive}
Maxim~V Shcherbakov, Artem~V Glotov, and Sergey~V Cheremisinov.
\newblock Proactive and predictive maintenance of cyber-physical systems.
\newblock In {\em Cyber-Physical Systems: Advances in Design \& Modelling},
  pages 263--278. Springer, 2020.

\bibitem{3}
Ovidiu Vermesan, Peter Friess, et~al.
\newblock {\em Internet of things-from research and innovation to market
  deployment}, volume~29.
\newblock River publishers Aalborg, 2014.

\bibitem{willner2017semantic}
Alexander Willner, Christian Diedrich, Ra{\'e}d~Ben Younes, Stephan Hohmann,
  and Andreas Kraft.
\newblock Semantic communication between components for smart factories based
  on onem2m.
\newblock In {\em 2017 22nd IEEE International Conference on Emerging
  Technologies and Factory Automation (ETFA)}, pages 1--8. IEEE, 2017.

\end{thebibliography}
